# A Survey on Instance Segmentation: State of the art

Abdul Mueed Hafiz[1*], Ghulam Mohiuddin Bhat[2]

[1, 2] Department of Electronics and Communication Engineering,
Institute of Technology, University of Kashmir,
Srinagar, J&K, India, 190006.

Corresponding Author Email[1*]: mueedhafiz@uok.edu.in
Co-author Email[2]: drgmbhat@uok.edu.in

ORC-ID[1]: 0000-0002-2266-3708
ORC-ID[2]: 0000-0001-9106-4699

**Abstract**

Object detection or localization is an incremental step in progression from coarse to fine digital image inference. It not only provides the classes of the image objects, but also provides the location of the image objects which have been classified. The location is given in the form of bounding boxes or centroids. Semantic segmentation gives fine inference by predicting labels for every pixel in the input image. Each pixel is labelled according to the object class within which it is enclosed. Furthering this evolution, instance segmentation gives different labels for separate instances of objects belonging to the same class. Hence, instance segmentation may be defined as the technique of simultaneously solving the problem of object detection as well as that of semantic segmentation. In this survey paper on instance segmentation- its background, issues, techniques, evolution, popular datasets, related work up to the state of the art and future scope have been discussed. The paper provides valuable information for those who want to do research in the field of instance segmentation.

## 1. Introduction

### 1.1 Background

Semantic segmentation [1] vis-à-vis its relation to deep learning [2-7], can be understood by consideration of the fact that the former is not an isolated research area, rather an incremental research approach in the continuation of coarse inference towards fine inference. Its origins can be traced to automated classification approaches [8-21]. Classification in turn may be defined as the process of predicting a complete input, i.e., prediction of class of an object in the image or providing a list of classes of objects in an image according to their classification scores. Detection or object localization is an incremental step from coarse to fine inference, which provides not only classes of the image objects but also gives the location of the classified image objects in the form of bounding boxes or centroids. The goal of semantic segmentation is to obtain fine inference by predicting labels for each image pixel. Every pixel is class-labelled according to the object or region within which it is enclosed. Going in this direction, instance segmentation provides different labels for separate instances of objects belonging to the same object-class. Thus, instance segmentation can be defined as the task of finding simultaneous solution to object detection as well as semantic segmentation [22]. Part-based segmentation continues the evolution of this research by decomposing each of the segmented objects into their respective sub-components. Figure 1 depicts this image segmentation evolution. In this survey paper, we will focus on instance segmentation, its techniques, its frequently used datasets, related work and its potential scope for the future.

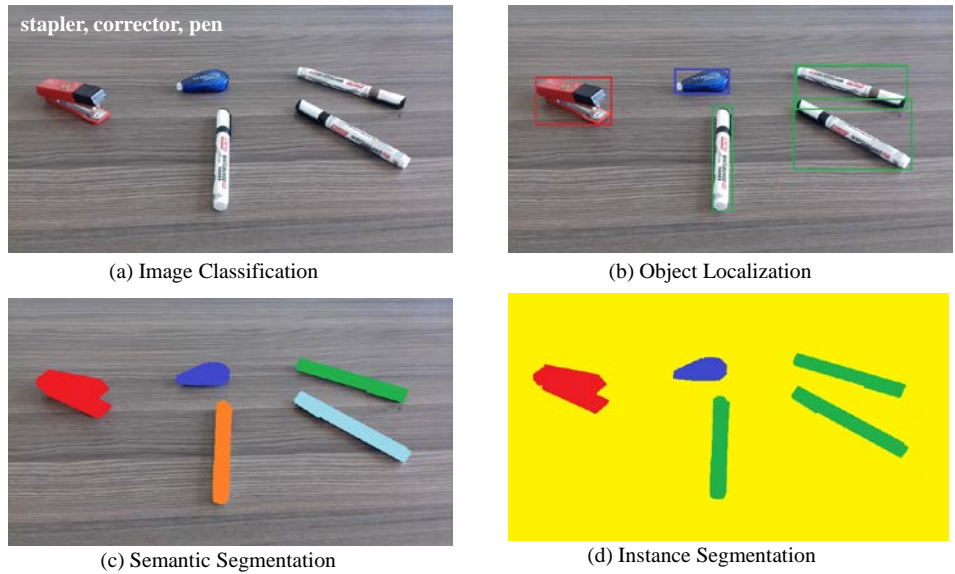

(a) Image Classification    (b) Object Localization

(c) Semantic Segmentation    (d) Instance Segmentation

**Figure 1**. Object recognition evolution: from coarse- to fine-grained inference: (a) Image Classification, (b) Object detection or localization, (c) semantic segmentation, (d) Instance segmentation.

### 1.2 Issues

The idea of semantic segmentation is developing a technique/algorithm that performs well in the two domains of *better segmentation accuracy* and *better segmentation efficiency*. Better segmentation accuracy encompasses accurate localization and recognition of objects in images/frames, with the result that the large variety of object related categories in real scenario can be distinguished (i.e. better distinctiveness), and that instances of objects belonging to same class which are subject to intra-class appearance variation, may be localized and recognized (i.e. better robustness). Better segmentation efficiency refers to computational cost of the segmentation algorithm. It refers efficient real time computational costs like acceptable memory/storage requirements, and lesser burden on the processor(s).

One of the important components in an object detector for segmentation is good feature representation which is of primary importance in object detection [23-27]. Previously, a lot of effort was put in designing local descriptors (like SIFT [28] and HOG [29]) and in exploring approaches (like Bag of Words ([30] and Fisher Vector [31]) in order to group and to abstract descriptors into high level representations for emerging the discriminative parts. The downside was that these feature representation methods needed handcrafted fine engineering and a large amount of domain expertise. As against this, methods based on deep learning (like Deep CNNs) are able to learn powerful representations of features with various abstraction levels from images [32,4]. Subsequently, the problem of feature representation has been transferred to the development of better performing network architectures and more optimized training procedures.

The trend in evolution of network architecture is of increasing depth: AlexNet [33] had eight layers, VGGNet [34] had 16 layers, and more recently ResNet [35] and DenseNet [36] both have more than 100 layers. In fact, it was VGGNet and GoogLeNet [37] which showed that with increasing the network depth, it is possible to increase the network's power of representation. Deep networks like AlexNet, OverFeat [38], ZFNet [39] and VGGNet have an extremely large number of parameters although they have few layers. This can be attributed to enormous number of parameters coming from the fully connected layers. Newer networks like Inception [40], ResNet, and DenseNet, although having a great depth, have far fewer parameters by avoiding the use of fully connected layers.

Deep CNN based detectors like RCNN [24], Fast RCNN [41], Faster RCNN [42] and YOLO [43], usually use the deep CNN architectures and subsequently use features from the topmost CNN layer for object representation. But there is a problem. Detection of objects across various scales is a big challenge. In order to address this issue, the detector is run over a pyramid of images [44,24,45]. Though this approach typically leads to more accurate detection, however, it suffers from limitations of inference time and also that of computational resources like memory.

Instance segmentation for small objects remains an issue. CNNs compute features in a layer by layer hierarchy, hence the sub-sampling layers in the hierarchy of features by default lead to an inbuilt multi-scale pyramid, and in turn produces maps of features at various resolutions. This behaviour leads to issues [46-48]. For example, higher CNN layers have a broad receptive field with more robustness to variations in pose, deformation and illumination, but resolution is lower and detail is lost. As against this, lower CNN layers have a narrow receptive field with richer detail, but resolution is higher and sensitivity to semantics in much lesser. Semantic attributes of objects emerge in various layers, which in turn depend on size of the objects. Hence if an object is small, its detail in earlier CNN layers is less, and the same can almost disappear in higher layers. This issue makes small object detection quite challenging. Various techniques have been proposed to tackle this issue e.g. dilated convolution [49-51], and increasing resolution of features. However these techniques lead to higher computational complexity. Also, if the object is large, then its semantic concept will be reflected in higher layers. Many techniques [48,52-54] have been developed for improving the detection accuracy by using various CNN layers.

Another issue is the handling of geometric transformations. DCNNs by nature cannot be spatially invariant with regards to geometric transformation [55-57]. Local max pooling in DCNN layers allows the networks to have some degree of translational invariance. In spite of this, the intermediate maps of features are not actually transformation invariant [55].

Handling of occlusions is also an issue. In real images, occlusions commonly occur which result in loss of information from instances of objects. For this problem, deformable ROI pooling [58-60] and deformable convolution [58] were proposed. These techniques alleviate occlusions by making fixed geometric structures more flexible. Wang et al. [61] also proposed training of an adversarial network [62]. In spite of these efforts, the problem of occlusions has not been even merely solved. Using GANs in order to address this problem looks promising.

Handling of image degradations is also an issue. Noise in real-world images is a problem. This is usually caused by problems in lighting, low quality in cameras, compression in images, etc. Although low quality in images tends to degrade their recognition, most contemporary techniques are benchmarked in a degradation free environment. This is justified by the fact that image databases like ImageNet [63], Microsoft COCO [64], PASCAL VOC [65], etc. all use high quality images. As of now it is observed that there is a miniscule body of work to address this issue.

**1.3    Instance Segmentation**

Instance segmentation has come to be one of the relatively important, complex and challenging areas in machine vision research. Aimed at predicting the object class-label and the pixel-specific object instance-mask, it localizes different classes of object instances present in various images. Instance segmentation aims to help largely robotics, autonomous driving, surveillance, etc.

With the advent of deep learning [5,4] more specifically Convolutional Neural Networks (CNNs) [33,66,39], many instance segmentation frameworks were proposed, for example [67-69,6], in which the segmentation accuracy grew rapidly [65]. Mask R-CNN [67] is a straightforward and efficient instance segmentation approach. Taking a lead from Fast/Faster R-CNN [41,42], a fully convolutional network (FCN) has been used to predict segmentation masks, side by side with box-regression and object classification. For high performance, feature pyramid network (FPN) [53] has been used to extract stage-wise network features, in which a top-down network path having lateral connections has been used to obtain features which are semantically strong.

Some relatively new datasets provide adequate room for improvement of proposed techniques. Microsoft's Common Objects in Context or COCO dataset [64] has 200k images. Many instances having complicated spatial layout have been captured in the images of this dataset. Also, the Cityscapes dataset [70] and the Mapillary Vistas Dataset or MVD [71] have street scene images containing a large number of traffic objects per image. Blurring, occlusion and minute instances are found in the images of these datasets.
Many principles have been proposed for network design for classifying images. The same are also substantially useful for object recognition. Some examples in this regard are shortening of information path [35,72], using dense connections [36], increasing information-path flexibility and diversity by creation of parallel paths [73,74], etc.

From the 100+ papers covered in this survey, there are 4 top level research clusters which we describe next.

## 2.  Instance Segmentation Techniques: A Taxonomy

**2.1    Classification of mask proposals**

Figure 2 shows the general framework for this class of techniques.

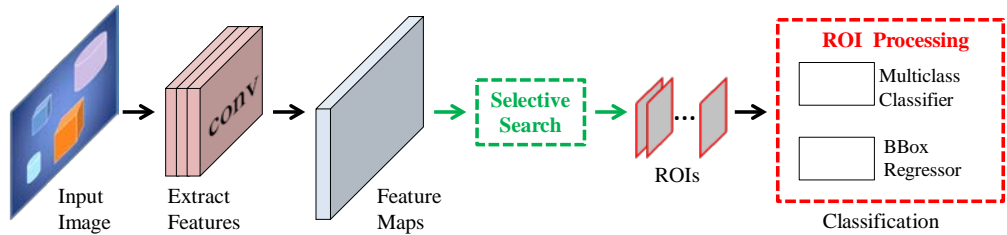

**Figure 2.** General framework for *Classification of Mask Proposals* Techniques

2.1.1 Bottom-up Mask Proposals

Before being popularized by COCO [64], instance segmentation, in its modern sense, was introduced by Hariharan et al. [75]. The proposed technique involved generation of mask proposals [76,77], followed by classification of the generated proposals [75]. Earlier, this classification of mask proposals approach was used elsewhere. As a example, Selective Search [76] used this technique for obtaining box detections and performing semantic segmentation. These techniques could conveniently be used for instance segmentation.

2.1.2 Deep Learning

The prior techniques depended on doing bottom-up mask proposal generation, before deep-learning became popular [76,77]. Subsequently, the former were replaced by new techniques having a more efficient structure, which came along with such as RCNN [24]. In spite if their better segmentation accuracy, RCNN and the other techniques inside this band, suffered from some issues. For example, training was based on a multistage pipeline, which was slow and was difficult to optimize, due to the need to train each stage separately. Features had to be extracted for each proposal in every image from the CNN, leading to storage, time and detection-scale issues respectively. Testing was also slow due to the need to extract CNN features. Subsequently, RCNN was followed by Fast RCNN [41] and Faster RCNN [42], which addressed its problems.

## 2.2 Detection followed by segmentation

The popular approach for instance segmentation involves object detection using a box followed by object-box segmentation [78,72,68,67]. Figure 3 shows the general framework for this class of techniques.

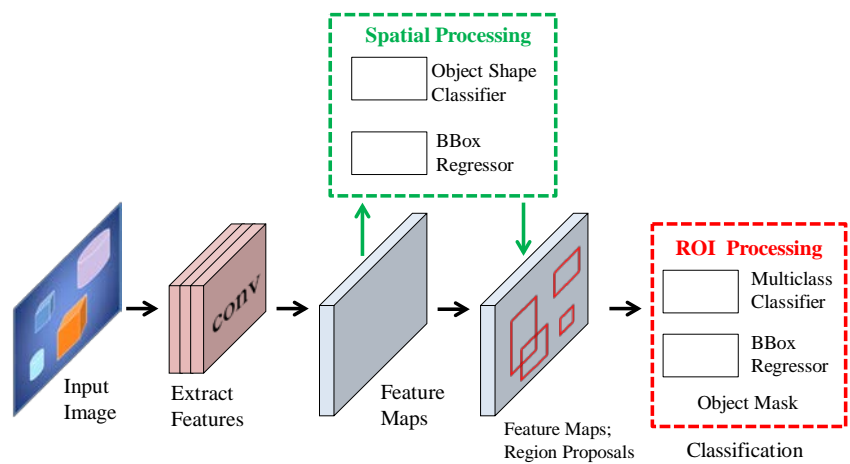

**Figure 3.** General framework for *Detection Followed by Segmentation* Techniques

### 2.2.1 Mask Based Techniques

One of the most successful techniques in this regard can be Mask RCNN [67]. Mask RCNN extends the Faster R-CNN [42] detection algorithm using a relatively simple mask predictor. Mask RCNN is simple to train, has better generalization, and only adds a small computational overhead to Faster R-CNN. The former runs at 5 FPS. Instance segmentation approaches that bank on Mask R-CNN [79-81] have shown promising results in recent instance segmentation challenges [64,71,70].

### 2.2.2 Other Techniques

For the purpose of detecting object bounding boxes, the following techniques have been used.

- sliding-window techniques [82-84]
- region-based techniques [41,42]

In spite of their strengths of the techniques discussed in Sections 2.2.1 and 2.2.2, they depend on pipelining leading partially to some of the drawbacks discussed in Section 2.1.

## 2.3 Labelling pixels followed by clustering

Another approach to instance segmentation e.g., [69,85,86], uses techniques created for the task of semantic segmentation [7,87]. This approach involves categorical labelling of every image pixel. This is followed by grouping pixels into object instances using a clustering algorithm. Figure 4 shows the general framework.

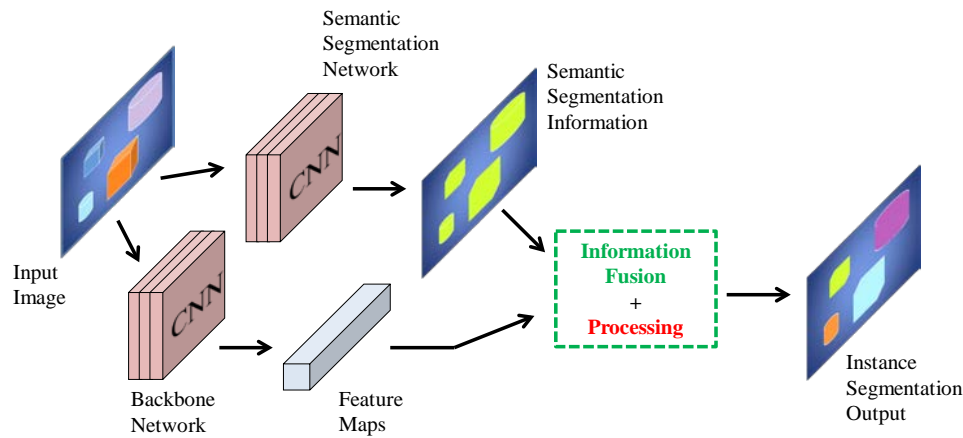

**Figure 4.** General framework for *Labelling Pixels Followed by Clustering* Techniques

The approach benefits from recent positive developments in semantic segmentation which can predict a high resolution object mask. In comparison to detection-followed-by-segmentation techniques, labelling-pixels-followed-by-clustering methods have lesser accuracy on frequently used benchmarks [64,71,70]. Due to intense computation necessitated by pixel labelling, more computational power is generally required.

## 2.4 Dense sliding window methods

The general framework for this class of techniques is shown in Figure 5.

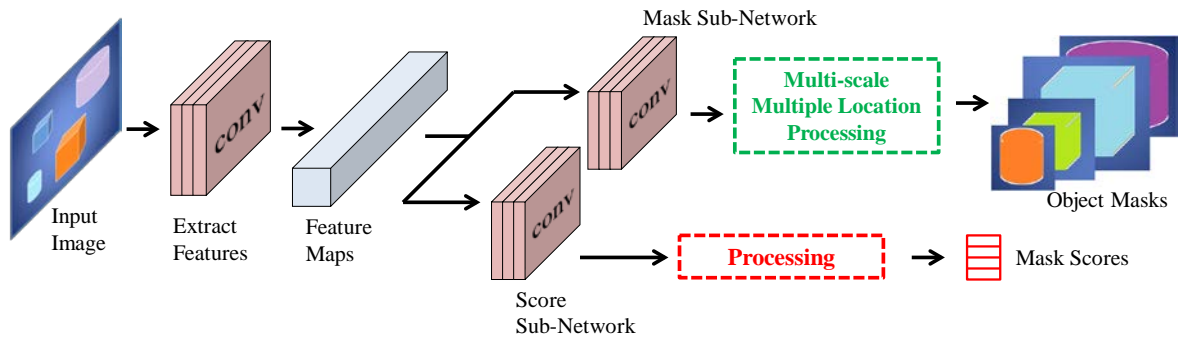

**Figure 5.** General framework for *Dense Sliding Window Methods*

2.4.1   Class Agnostic Mask Generation Techniques

These techniques e.g. DeepMask [88,89], InstanceFCN [90], etc. use CNNs for mask proposal generation by dense sliding-window techniques.

2.4.2   TensorMask

TensorMask [91] uses a different architecture as compared to the techniques discussed in Section 2.4.1 leading to better performance. Also, unlike DeepMask and InstanceFCN, TensorMask involves classification for multiple classes, which is done in parallel with predicting masks. This feature makes it useful for instance segmentation.

The techniques (discussed in Section 2.4.1 and 2.4.2) e.g. TensorMask, have a decent performance on benchmarking datasets like COCO, however the algorithmic complexity is an issue.

Tabular taxonomy of the notable methods discussed in Section 2 is given in Table 1.

**Table 1**. Tabular taxonomy of the notable techniques mentioned in Section 2

| Group | Technique |
| --- | --- |
| **Classification of mask proposals** | RCNN, Fast RCNN, Faster RCNN |
| **Detection followed by segmentation** | HTC, PANet, Mask RCNN, Mask Scoring RCNN, MPN, YOLACT |
| **Labelling pixels followed by clustering** | Deep Watershed Transform, Instance Cut |
| **Dense sliding window methods** | Deep Mask, Instance FCN, Tensor Mask |

## 3. The Evolution of Instance Segmentation

Although there is a large number of techniques applied to instance segmentation, the notable methods in light of the discussion from Section 2 are discussed below which are also shown on the timeline in Figure 6.

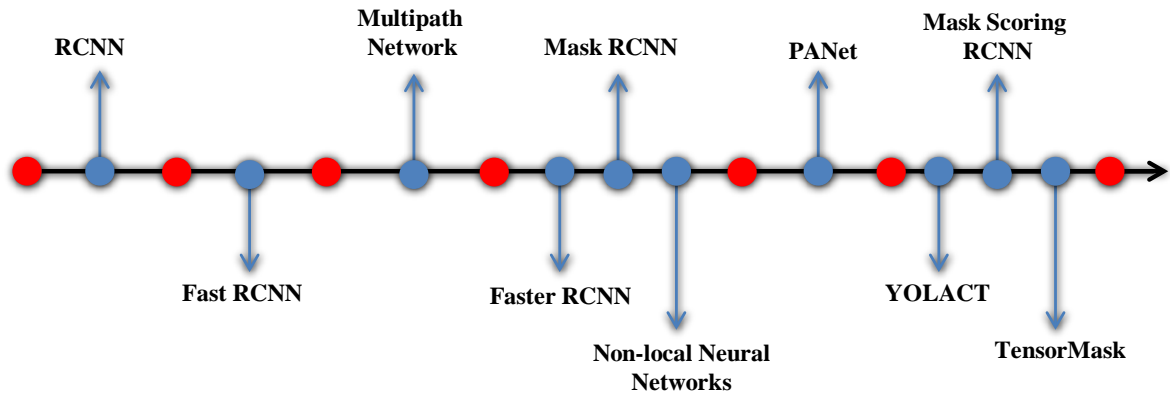

Figure 6. Timeline for notable techniques in instance segmentation

### 3.1 RCNN

After being inspired by the breakthroughs in image classification, obtained by use of CNNs and the success of selective search technique in region proposals for manually generated features [76], Girshick et al. [24] were one of the first to explore CNNs for instance segmentation [27]. They developed the RCNN technique which integrated AlexNet [33] along with a region proposal using the selective search technique [76]. Training an RCNN model consists of following steps. First step involves computing class agnostic region proposals obtained using selective search. Next step is CNN model fine-tuning which consists of using the region proposals for fine-tuning a pre-trained CNN model like AlexNet. Next, a set of class specific Support Vector Machine (SVM) classifiers are trained on features extracted from the CNN which replace the softmax classifier learned by fine-tuning. This is followed by class specific bounding box regressor training for each object class using features obtained from the CNN.

Although RCNN achieved high object detection quality, it has some notable drawbacks. For example, training in a multistage pipeline is slow and difficult because each individual stage has to be trained separately. Also, for training the SVM classifier and BBox regressor respectively, more resources and time are needed. Finally, testing is slow, because features from CNN have to be extracted for each object proposal in every testing image sans shared computation.

These problems with RCNN inspired development of other techniques, which led to birth of improved detection frameworks e.g. Fast RCNN, and Faster RCNN.

### 3.2 Fast RCNN

Fast RCNN [41] addressed some of the issues of RCNN, and subsequently improved its object detection ability [27]. Fast RCNN uses end-to-end training of the detector. It does this by streamlining the training process by simultaneous learning of the softmax classifier and of the class specific BBox regression, rather than individually training various components of the model as done in RCNN. Fast RCNN shares the computation of convolution among region proposals, and subsequently adds an ROI pooling layer between the last convolution layer and the first fully connected layer in order to extract features for every region proposal. ROI pooling uses the concept of feature level warping in order to achieve image level warping. The ROI pooling layer features are given into a sequence of fully connected layers which finally branch into 2 layers viz. object category prediction softmax probability, and class proposal refinement offsets. In comparison to RCNN, Fast RCNN improves the efficiency to a large extent e.g. by 3 times training speed and by 10 times in testing speed.

### 3.3 MultiPath Network

In the MultiPath Network technique [72], three modifications have been done to the standard Fast R-CNN model. First, skip connections [92,47,93] have been incorporated which give the object detector access to features of different network layers. Second, a foveal element has been added for exploitation of context of objects at different resolutions. And finally, a loss function of integral nature has been added. The network has been adjusted in order to improve localization of the instance segmentation masks. As a result of the modifications mentioned, a multiple-path information flow gets created in the network, hence the name, "MultiPath" network. The MultiPath network has been coupled with DeepMask model, the latter being quite convenient for localizing object even if they are small. The pipeline has also been adapted for predicting masks as well as bounding boxes. The new model demonstrated improvement over Fast R-CNN model [41] in overall by 66% and by four times on small object instances. It ranked second in both 2015 detection and segmentation challenges for the COCO dataset.

### 3.4 Faster RCNN

In spite of the fact that Fast RCNN led to significant speed up in detection, it still relied on external region proposals, for which computation was the speed bottleneck in Fast RCNN [27]. At that point in time, work [94,95] showed that CNNs have the ability of object localization in convolutional layers, an ability which is weakened in fully connected layers. Hence it was found to be feasible that selective search could be replaced by a CNN for production of region proposals. Subsequently, Faster RCNN model was proposed by Ren et al. [42] which had a Region Proposal Network (RPN) for generation of region proposals and this was efficient and accurate. The same backbone network was used, taking features from the last shared convolutional layer in order to accomplish region proposal by RPN and region classification by Fast RCNN.

### 3.5 Mask R-CNN

In [67], the authors present Mask R-CNN, a relatively simple and flexible model for instance segmentation. The model conveniently performs instance segmentation by object detection with simultaneous generation of high quality masks. Mask R-CNN furthers the Faster R-CNN [42]. Normally, Faster R-CNN has a branch for object bounding box recognition. Mask R-CNN adds an object mask prediction branch in parallel as an improvement. The head architecture using an FPN backbone is shown in Figure 7.

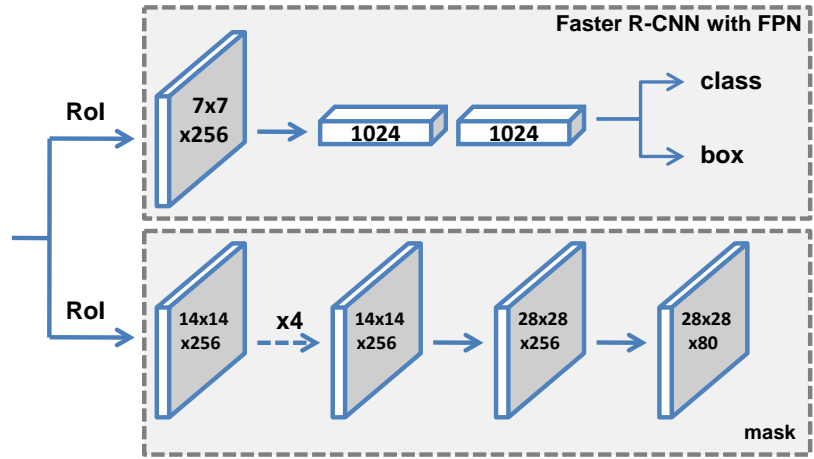

**Figure 7**. Head Architecture: Extension of existing Faster RCNN head with FPN backbone to which a mask branch is added. Numbers denote spatial resolutions and channels. Arrows denote convolution, deconvolution, or fully connected layers as can be inferred from context [convolution preserves spatial dimension while deconvolution increases it). ([67])

The proposed model is relatively easy to train, while adding a small computational load to the Faster R-CNN model which runs at 5 fps. Another advantage of Mask R-CNN is ease of generalization with regards to other related tasks. For example, Mask R-CNN allows estimation of human poses in a similar environment. Mask R-CNN achieved first position in all the 3 COCO suite 2016 challenges viz. instance-segmentation, detection of bounding boxes, detection of person key points. The model, as demonstrated by the authors, outperformed other state of the art models on each COCO task for the 2016 challenge of the dataset.

### 3.6   MaskLab

MaskLab [22] improves Faster R-CNN [42] and produces 2 additional outputs viz. semantic segmentation and instance center direction [96]. The prediction boxes given by Faster R-CNN bring the instances of the objects having different scales to a canonical scale, and then MaskLab does foreground and background segmentation inside each prediction box by using both semantic segmentation as well as direction prediction. The semantic segmentation prediction by encoding the pixel-wise classification data along with background class has been adopted in order to distinguish between objects of various semantic classes. This technique removes the duplicate background encoding in [68]. In addition to this, the direction prediction has been used for separation of instances of an object with common semantic label. MaskLab employs same assembling technique as used in [90,68] for collection of the direction information to get rid of the complicated template matching technique used in [96]. In addition to this, by taking motivation from the recent advances in both segmentation as well as detection, the proposed model further incorporates atrous convolution [97] for extraction of denser features maps, hypercolumn features [92] for the purpose of refining mask segmentation [98], multi-grid technique [99,58,100] for capturing various context scales, and a new TensorFlow technique [101], deformable cropping and resize, having been inspired by deformable pooling operation [58]. The performance of their model is comparable to other state of the art models.

### 3.7   Non-local Neural Networks

Non-local means [4] is a filtering technique which computes a weighted mean of all pixels in an image. In doing so, it allows distant pixels to contribute to the filtering response at a location which is based on path appearance similarity. This idea was successively developed by Block-matching 3D (BM3D) [102-105]. Long-range dependencies have been modelled by graphical models e.g. conditional random fields (CRF) [106,107]. The mean-field inference in a CRF can be converted into a recurrent network and subsequently can be trained [108-112]. The authors of this work [113] claim that their technique is simpler and the collective work of theirs and others are related to graph neural networks [114]. They further claim that their work is related to the *self-attention* [115] method used in machine translation. A self-attention capsule calculates the response at a position in a sequence e.g. a sentence, by looking at all positions and subsequently taking their weighted average inside an embedding space. Self-attention can be viewed as a non-local mean [116], and hence can thus their work connects self-attention in machine translation with the general class of non-local filtering operations applicable to image and video problems in machine vision. The authors claim that they address non-local modelling, which is a long-time important component of image processing [117,116], has been overlooked to a large extent in recent neural networks for machine vision and that they address this issue.

It is notable that convolution and recurrent operations, both, are the building blocks which process a single spatially local neighbourhood at a given time. The authors of this approach [113] propose a family of non-locally operating building blocks for the purpose of capturing long range pixel dependencies. The non-local mean operation [116], can be defined as a generic non-local operation in deep neural networks as:

$$y_i = \frac{1}{C(x)} \sum_{\forall j} f(x_i, x_j) g(x_j) \qquad (6)$$

Here $i$ is the index of the output position whose response is to be calculated and $j$ is the index which gives all possible positions. $x$ is the input signal and $y$ is the output signal which has same size as $x$. $f$ is a pair-wise function which computes a scalar value like affinity between $i$ and all values of $j$. $g$ is a unary operation which computes the representation of the signal at the input at position $j$. The response is normalized by $C(x)$. The authors claim that a non-local operation is a flexible building block which can be conveniently used with convolutional and recurrent neural networks, unlike fully connected layers that are usually used in the end of these networks. They have used this operation and built a rich architecture which is able to combine non-local and local information. Figure 8 shows a spacetime non-local block.

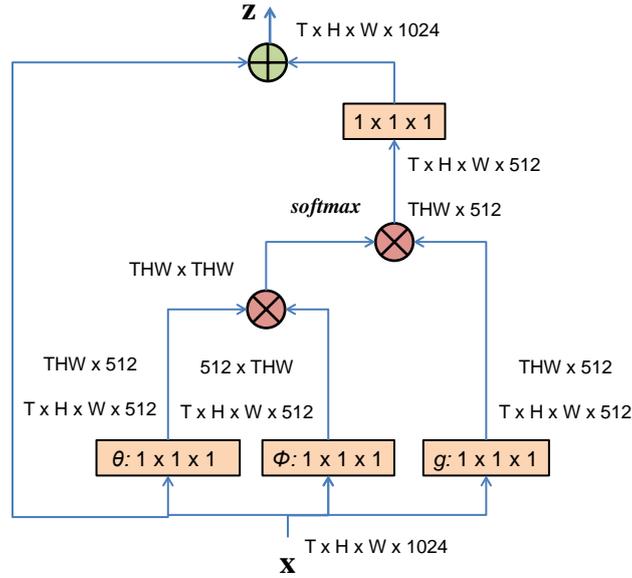

**Figure 8**. A spacetime non-local block. Feature maps are denominated by the shape of the tensors, for example TxHxWx1024 is for 1024 channels. Matrix multiplication, and element-wise sum are denoted by red circles and green circles respectively. The softmax operation is performed on every row. The orange boxes denote 1x1x1 convolution. An embedded Gaussian version with a bottleneck of 512 channels is shown above. ([113])

For the purpose of classification of video frames, the authors claim that the non-locally operating models are competent or even capable of outperforming the state of the art approaches used on the Kinetics and the Charades datasets. The technique has been demonstrated to improve instance segmentation on COCO dataset.

### 3.8 Path Aggregation Network (PANet)

The authors of [79] discuss that although the practice of using features from different layers in image recognition is well established [89,118,119,47,120,121,48,53,82,122,123]. Accordingly they take FPN [53] as baseline and enhance it significantly. They also discuss that some research groups [93,124,125,92] concatenated feature grids from various layers for higher efficiency. But they note that a sequence of operations like normalization, concatenation and dimension reduction are required to get useful new features. They claim that their design is much simpler while giving good results. Feature grid feature fusion has been used previously with feature maps on input with different scales [126]. The authors note that their technique uses single-scale input. End to end training is used. The authors note that some research groups [25,127,124] have used techniques of pooling features for every proposal with a foveal structure for exploitation of contextual information from regions with different resolutions e.g. global pooling [128,129,118]. The authors note that they have used the mask prediction branch which also supports accessing global information in a novel way.

The framework of the proposed technique is shown in Figure 9. The authors of PANet note that manner in which information flows or propagates in deep neural networks is very important. The authors of PANet Model [79] aim at boosting the flow of information in the proposal based framework used for instance segmentation task. They improve the deep network hierarchy of features with specific signals related to localization in the lower layers. This process is referred to as bottom up path-augmentation. It leads to shorter information paths between the lower layers and the features at the top of the deep network. They also propose a technique referred to as adaptive feature-pooling which relates the grid of features and features at all levels. Due to this technique, relevant information in every level of features flows to the subsequent sub-networks used for generating proposals. An alternate branched segment captures various proposal views in order to enhance the prediction of the generated masks.

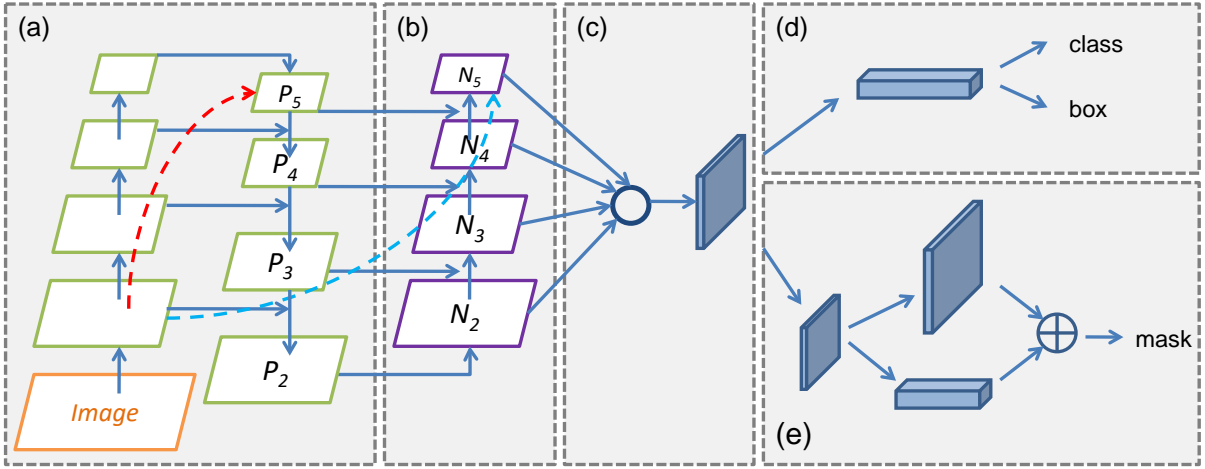

**Figure 9.** PANet framework. (a) FPN backbone. (b) Bottom-up path augmentation. (c) Adaptive feature pooling. (d) Box branch. (e) Fully-connected fusion. ([79])

The techniques proposed are relatively easy to implement and have relatively small overhead with regards to computational load. PANet achieved the first position in the 2017 Instance Segmentation Challenge Task for COCO and also achieved the second position in the task of object detection sans batch-training for large number of images.

### 3.9 Hybrid Task Cascade

Cascade combination of models has proven to be a useful approach for boosting performance of different tasks. Introducing cascading to instance segmentation is a challenging area. In this direction, the authors introduce Hybrid Task Cascade [81]. The authors begin with direct combination of Mask R-CNN and Cascade R-CNN which they call Cascade Mask R-CNN. Specifically, a mask branch on lines of the Mask R-CNN is added to each stage of Cascade R-CNN. The pipeline [81] is given by:

$$x_t^{box} = P(x, r_{t-1}), \qquad r_t = B_t(x_t^{box}),$$

$$x_t^{mask} = P(x, r_{t-1}), \quad m_t = M_t(x_t^{mask}) \qquad (1)$$

In the above equation, x is the CNN feature of the network, $x_t^{box}$ and $x_t^{mask}$ indicate box and mask features obtained from x and the input region-of-interests (ROIs). P(·) is the pooling operator, for example ROI align. $B_t$ and $M_t$ refer to the box head and mask head at the $t^{th}$ stage, $r_t$ and $m_t$ refer to the corresponding box and mask predictions respectively. By combination of the advantages of cascaded refinement, and the benefits from bounding boxes and mask predictions, the technique improves the box AP, in comparison to Mask R-CNN and Cascade R-CNN. But the authors note that the performance is unsatisfying. The authors propose interleaving of the box branch and mask branch. This execution is expressed as:

$$x_t^{box} = P(x, r_{t-1}), \qquad r_t = B_t(x_t^{box}),$$

$$x_t^{mask} = P(x, r_t), \qquad m_t = M_t(x_t^{mask}) \qquad (2)$$

The authors note that, by doing the above the mask branch is able to take advantage of the updated predictions of the bounding boxes. The authors next introduce information flow among the mask branches by manner of feeding the masks of the previous stages to the new stages.

$$x_t^{box} = P(x, r_{t-1}), \qquad r_t = B_t(x_t^{box}),$$

$$x_t^{mask} = P(x, r_t), \qquad m_t = M_t\big(F(x_t^{mask}, m_{t-1}^-)\big) \qquad (3)$$

Where $m_{t-1}^-$ refers to the intermediate features of $M_{t-1}$ which is used as the representation of the mask at stage $t - 1$. F is the combination function. The authors propose the following implementation, wherein they adopt ROI features prior to the deconvolution layer as the mask representation $m_{t-1}^-$ with spatial size 14x14.

$$F(x_t^{mask}, m_{t-1}) = x_t^{mask} + G_t(m_{t-1}^-) \qquad (4)$$

A notable contribution of this work is the use of spatial contexts by addition of a new branch for prediction of per-pixel semantic segmentation for the entire image. Figure 10 shows the architecture of this branch.

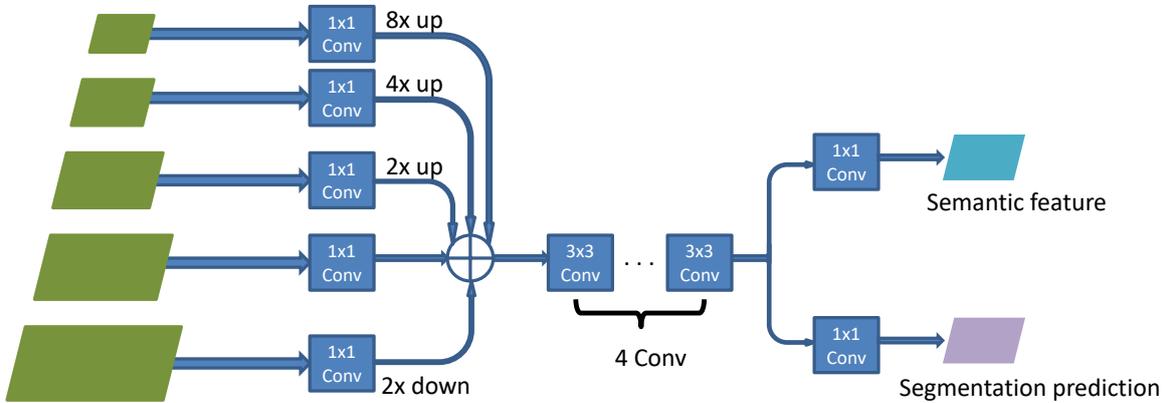

**Figure 10.** Architecture of the Semantic Segmentation Branch ([81])

The authors of the work discuss that they found the key to optimum cascading for instance segmentation was to take maximum advantage of the inverse relationship between object detection and object instance segmentation. Hybrid Task Cascade or HTC differs from conventional cascading in 2 important ways. First, instead of using refined cascading on the two tasks, HTC processes them in multiple stages in a combined manner. Second, it uses a fully convolutional segment for providing spatial context. This helps in distinguishing foreground from noisy background. The authors claim that HTC is able to learn more useful features by integration of features which are complementary, progressively with every stage. Without any fine tuning, an HTC model obtained 38.4% mask AP with 1.5% improvement

w.r.t. a cascaded Mask R-CNN for the COCO dataset. Further, the proposed model achieved a score of 48.6 Mask AP on the test-challenge subset of COCO, achieving first position for the COCO 2018 Challenge in object detection.

### 3.10 GCNet

The authors of Global Context Network (GCNet) [130] note that the Non-Local Networks [113] (discussed in the original work given in Section 3.7) present a novel approach for capturing long-range dependencies by aggregation of query-specific global context for every query location in the image. In spite of this, they found through a thorough mathematical analysis that global contexts which are modelled by non-local networks are almost the same for various query positions throughout the image. The authors claim to have taken advantage of this finding in order to create a simple network which is based on query-independent formulation. They claim that the proposed network maintains the accuracy of Non-local Networks but with much lesser computational expenditure. The authors of GCNet note that their design is similar in structure to Squeeze-Excitation Network (SENet) [131]. They propose unification into a three-step general model for modelling of global context. Inside the general model, a more efficient instantiation, referred to as the Global Context (GC) Block has been designed. The block is lightweight and is able to efficiently model the global context. The fact that it is lightweight allows the designers to apply it among multiple layers in the network, thus constructing a Global Context Network or GCNet. The GC Block is shown in Figure 11. It is formulated as:

$$z_i = x_i + W_{v2}\text{ReLU}\left(\text{LN}\left(W_{v1}\sum_{j=1}^{N_p}\frac{e^{W_k x_j}}{\sum_{m=1}^{N_p}e^{W_k x_m}}x_j\right)\right) \quad (5)$$

Where $x_i$ = input for each query position $i$,

$W_k$ and $W_v$ denote linear transformation matrices,

$\alpha_j = \frac{e^{W_k x_j}}{\sum_m e^{W_k x_j}}$ is the weight for *global attention* (mentioned in Section 3.7) *pooling,* and $\delta(\cdot) = W_{v2}\text{ReLU}(\text{LN}(W_{v1}(\cdot)))$ gives the bottleneck transform.

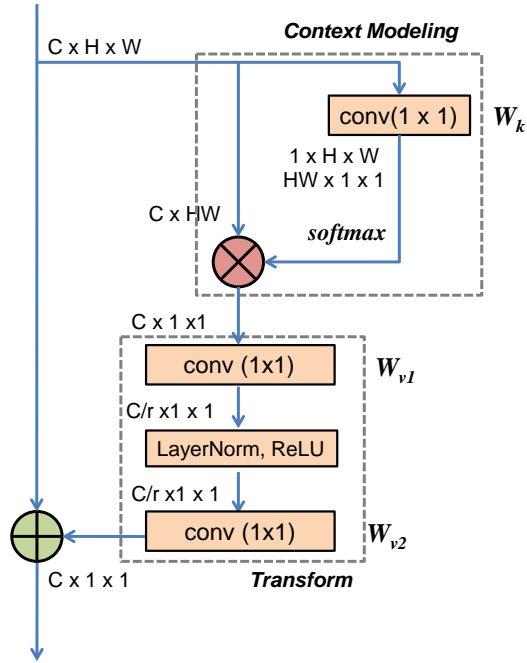

**Figure 11.** Architecture of the Global Context Block. Feature-maps have been shown as feature-dimensions. For example C x H x W denotes the feature-map with channel number C, with height H and with width W, respectively. Matrix multiplication, and element-wise sum are denoted by red circles and green circles respectively. (The architecture draws inspiration from the Non-Local Block given in Figure 8, whose original work is discussed in Section 3.7). 'r' is bottleneck ratio and C/r represents hidden representation dimension of the bottleneck. Default reduction ratio is set to r=16. ([130])

GCNet outperforms both Non-Local Networks and SENet on MS COCO.

### 3.11 YOLACT

YOLACT [132] is a fast and simple instance segmentation model with fully-convolutional topology. It is used for instance segmentation in real time and had the distinction of being the fastest real-time instance segmentation technique when it was introduced. It achieved a segmentation score of 29.8 Mask AP on the COCO dataset at 33 frames-per-second when one Titan XP GPU was used. This was faster than other state of the art approaches at that time. The experimentation used a single GPU for training. This result was achieved by first, bifurcation of image segmentation in parallel into 2 subtasks: (i) generation of prototype masks, and (ii) prediction of mask coefficients for each instance-mask. Next, instance masks are produced by linear combination of the prototype masks with the coefficients of the masks. The authors also analyzed the behaviour of emerging prototype masks and demonstrated that the network learnt localization of the object instances automatically with translational variance. This was in spite of the fact that the model was fully convolutional.

### 3.12 Mask Scoring R-CNN

An important task in the context of image segmentation is allowing a deep neural network to be aware of its prediction quality. For the purpose of instance segmentation, the confidence estimate of classification of the instance has been used as quality score of the mask by a majority of instance segmentation approaches. In contradiction to this approach, the quality of the mask, as quantified as the Intersection over Union or IoU between the mask of the instance and the ground truth, has not been related properly to the classification score. In [133], the authors have studied this problem and subsequently proposed Mask Scoring R-CNN. The authors of this work note that without losing generality, they have worked on Mask R-CNN [67] and added an additional MaskIoU head module that learns the Mask-IoU aligned mask score. They claim that their technique is conceptually simple. Mask R-CNN has been combined with MaskIoU Head, which feeds on the instance features and predicted mask in combination. This arrangement is used to predict the IoU between the input mask and the ground truth mask. Figure 12 shows the Mask Scoring R-CNN architecture.

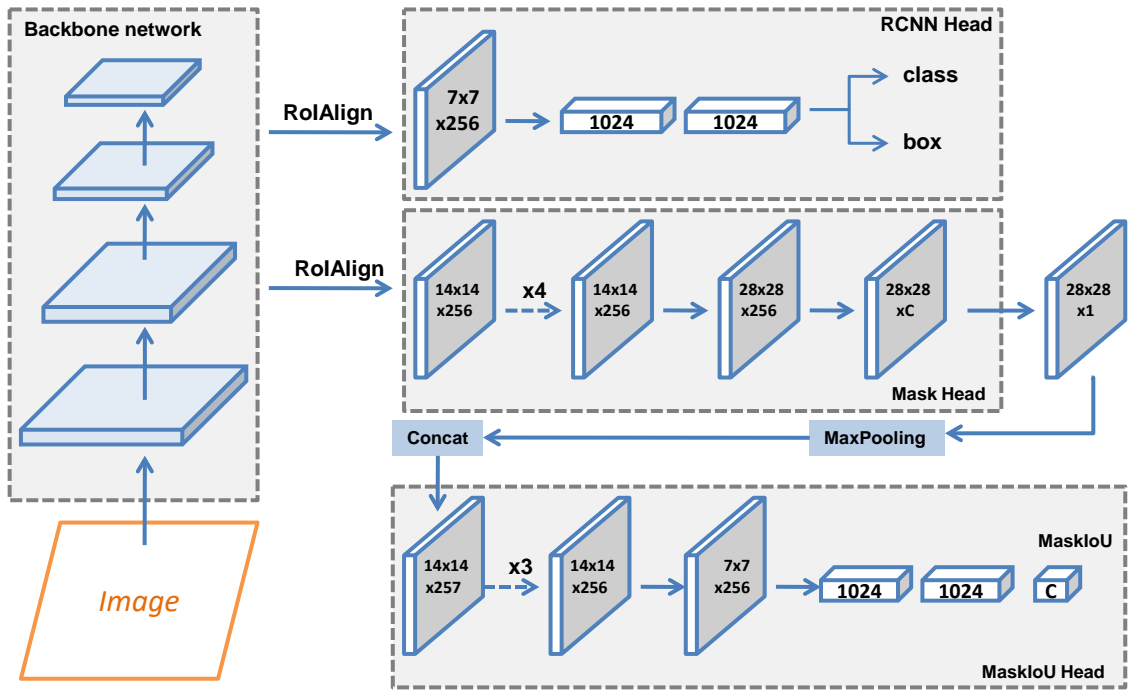

**Figure 12**. Architecture of Mask Scoring R-CNN model. Image is fed to the backbone network to generate the ROIs via R.P.N. and the ROI features via ROIAlign. RCNN Head and Mask Head are components of Mask R-CNN model. For predicting MaskIoU, they use the predicted mask and ROI features as input. MaskIoU Head has four Convolutional layers and have three fully connected layers. The last fully connected layer gives outputs for C classes MaskIoU. ([133])

The model contains a network block which has the ability to learn predicted mask quality. This network block proposed by the authors, combines the features of the instance with the predicted mask, in order to regress the predicted mask IoU. This mask scoring approach calculates the alignment error between quality of the mask and the score of the mask, thus improving the performance of the instance segmentation task by prioritizing better predictions of masks in COCO AP evaluation. Following extensive experimentation using the COCO dataset, the proposed approach consistently and noticeably improves various models. It even outperforms the efficient Mask R-CNN approach.

### 3.13 TensorMask

Object detection models using sliding window technique which generate predictions of bounding-box with the help of a dense and regularly spaces grid have shown rapid advancement besides garnering significant popularity. In [91], the authors propose a model referred to as TensorMask, wherein they perform instance segmentation using dense sliding windows. This is an area which is relatively unexplored. The output for every pixel location tends to be a geometric structure having its own dimensions w.r.t. spatial aspects. In this work, a dense form of instance segmentation is performed by prediction over four-dimensional tensors. The geometry of the same is captured and novel operators over four-dimensional tensors are used. The key concept of the TensorMask representation is using structured 4-D tensors for representation of masks over a spatial domain. This perspective is different from earlier work on the task of instance segmentation such as DeepMask [88] as well as InstanceFCN [90] which used unstructured 3-D tensors, wherein the segmentation mask has been packed in the third channel axis. Using a simple channel representation leads to loss of an opportunity to benefit from using structural arrays for representation of masks as 2-D entities. To overcome this problem, the authors of TensorMask propose to use 4-D tensors with shape $(V, U, H, W)$, wherein both $(H, W)$ for object position, and $(V, U)$ for representing relative mask position, are subtensors. By virtue of the TensorMask framework, the authors have developed a pyramidal structure on top of a scale-indexed list of the 4-D tensors called a tensor bipyramid. This structure has a pyramid shape in both $(H, W)$ and $(V, U)$ sub-tensors, however they grow in oppositely. They have combined these components into a network backbone and have followed the training procedure of RetinaNet [53] closely in which their dense mask predictors extend the original dense bounding box (BBox) predictors. The authors through their work on TensorMask stress on the importance of specifically capturing the geometric structure of this task. They show that TensorMask gives similar results to its Mask R-CNN counterpart. They claim that their results are promising and that the results suggest that the proposed framework/model can pave the way for future work on dense sliding window instance-based segmentation.

## 4. Datasets

A 2D image dataset contains gray-scale or RGB images. The information content is thus limited to spatial location of pixels and their intensity values. 2D image datasets are abundant for instance segmentation research. In the following section, we describe some of the most popular large 2D image datasets for instance segmentation.

### 4.1 Microsoft Common Objects in Context (COCO) Dataset

The Microsoft Common Objects in Context or COCO Dataset [64] is a large-scale image dataset for the purpose of recognizing, segmenting and captioning images. It has many challenges. The detection challenge is most relevant for instance segmentation. The detection challenge features above 80 object classes while providing above 82783 training images, 40504 validation images, and above 80000 testing images. The testing image set is divided amongst 4 subsets. These are: (i) test-dev with 20000 images for the purpose of additional validation and debugging, (ii) test-standard with 20000 images which is the default testing image data subset for different competitions and for state-of-the-art benchmarking, (iii) test-challenge with 20000 images for evaluation server challenge submission, and (iv) test-reserve with 20000 images is used to avoid possible challenge overfitting. The importance and popularity of the COCO dataset has increased substantially since it first appeared, mainly due to its large size. The challenge results are made available on a yearly basis in a combined workshop organized on lines of the European Conference on Computer Vision (ECCV) alongside those of ImageNet. Table 2 shows the notable benchmarking techniques on MS COCO Dataset.

**Table 2.** Notable Instance Segmentation work on the Microsoft COCO dataset

| Method | Average Precision (AP) | Year |
|---|---|---|
| Hybrid Task Cascade [81] (With Extra Training Data) | 43.9% | 2019 |
| PANet [79] | 42.0% | 2018 |
| SOLOv2 [134] | 41.7% | 2020 |
| GCNet [130] | 41.5% | 2019 |
| BlendMask [135] | 41.3% | 2020 |
| SOLO [136] | 40.4% | 2019 |
| Non-local Neural Networks [113] | 40.3% | 2017 |
| Mask Scoring R-CNN [133] | 39.6% | 2019 |
| CenterMask [137] | 38.3% | 2019 |
| MaskLab+ [22] | 38.1% | 2017 |
| TensorMask [91] | 37.3% | 2019 |
| Mask R-CNN [67] | 37.1% | 2017 |
| PolarMask [138] | 32.9% | 2019 |
| YOLACT [132] | 29.8% | 2019 |
| MultiPath Network [72] | 25.0% | 2016 |

### 4.2 Cityscapes Dataset

Cityscapes dataset [70] is a large collection of urban-street-scene images. It focuses on semantic understanding of the street scene. The dataset provides semantic, instance-specific, and pixel-specific annotations. 30 object classes have been grouped into 8 categories relevant to urban scenes like flat surfaces, vehicles, people, sky, etc. Cityscapes Dataset consists of about 5000 images with fine annotation and 20000 with coarse annotation. The images for the dataset were captured in fifty cities in a span of several months during daytime with good weather. First it was video-recorded. As such the video-frames had to be hand-selected for having aspects like high number of object classes, different scenes, and different backgrounds.

### 4.3 The Mapillary Vistas Dataset (MVD)

The Mapillary Vistas Dataset (MVD) [71] is another large street scene image dataset. It contains 25000 annotated images with 66 classes. The annotation has been done using a dense, fine-grained manual style with the help of polygonal delineation for individually demarcating different objects. The dataset is five times bigger than that of Cityscapes for fine-annotation. It features images from across the world which have been captured during different weathers, seasons and daytimes. The dataset images have been captured using different devices like cell-phones, cameras, etc. with different photographers. The aim of developing the database is to further develop the state of the art research in understanding street scenes.

## 5. Summary and Discussion

In this section, we would like to discuss the key factors and issues which have emerged in instance segmentation based on deep learning, [27].

### 5.1 Detection frameworks: Two Stage versus Single Stage

Using the number of stages as a means of classification of the framework, two major categories emerge for detection frameworks, viz. Region based (2 stage) and unified framework (Single stage):

- For platforms having rich computational resources, two-stage frameworks produce better accuracies than their single-stage counterparts. In fact most winning techniques in famous challenges are usually two-stage frameworks. This is because their framework is flexible and more suitable for region based detection e.g. Mask RCNN [67].
- Single-stage detectors e.g. YOLO[83] are usually faster than their two-stage counterparts due to lack of pre-processing, light backbone network, lesser number of candidate regions, and use of fully convolutional detection sub-network. However, single stage frameworks poorly detect small objects, which is not the case for two-stage frameworks.

Many attempts have been made to increase accuracy and efficiency of detectors leading to convergence towards some crucial design choices:

- Fully Convolutional Framework
- Exploration of complementary information from correlated tasks e.g. in Mask R-CNN.
- Use of sliding windows
- Information fusing from various layers of the backbone

## 5.2 Backbone networks

Backbone networks are one of the important factors for performance due to their role as discriminators of object feature representations. Though deep backbones like ResNet [35], ResNeXt [73], etc. have been used successfully, however these are computationally expensive.

## 5.3 Improvement in robustness of representation of objects

Various factors like object size, lighting, background, blur, resolution, noise, etc. contribute to challenges in object recognition. The important techniques used to handle these challenges are discussed below.

### 5.3.1 Object Size/Scale

Variation in scale of objects especially small ones, poses challenges. The main strategies used are mentioned below.

- Use of image pyramids: This simple and efficient technique helps enlarge small objects and also shrinks larger ones. Though computational expensiveness is an issue, but the use of such techniques is common for obtaining better accuracy.
- Use of features from different convolutional layers for different resolutions [53].
- Up-scaling to better resolution in the network, for detection of small objects [139].

### 5.3.2 Occlusion, deformation and other factors

There are techniques to handle transformation, occlusions and deformation e.g., by use of a spatial transformer network. This technique uses regression to obtain a deformation area and then warps the features according to the deformation area [58]. Rotation invariance is important in real-world scenarios but is lesser attended to here due to the fact that popular benchmark segmentation-based datasets (like MS COCO) do not have large variations in rotation. Handling occlusion is well researched in other areas, however it is lesser attended to in the current area.

## 5.4 Detection proposals

Detection proposals significantly reduce the search-space for instance segmentation candidates. After the success of RPN [42], which was able to integrate generation of proposals and detection into a single framework, CNN based detection proposal frameworks have dominated region proposal.

## 5.5 Strengths and weaknesses with various Instance Segmentation Techniques

Table 3 lists the strengths and weaknesses with various techniques discussed earlier (in Section 2).

**Table 3**. Strengths and weaknesses of groups (of techniques) mentioned in Section 2

| Group | Strengths | Weaknesses |
|---|---|---|
| **Classification of mask proposals** | ✓ Relatively simple to implement;<br>✓ Modest segmentation accuracy; | o Slow and difficult to optimize training;<br>o Storage, time and detection-scale issues during training;<br>o Slow testing;<br>o Not suited for real time applications; |
| **Detection followed by segmentation** | ✓ Relatively simple to train;<br>✓ better generalization;<br>✓ relatively faster (e.g. YOLACT);<br>✓ good segmentation accuracy; | o Depend on a complicated training pipeline which is difficult to train, and to optimize; |
| **Labelling pixels followed by clustering** | ✓ Use some recently investigated techniques;<br>✓ Relatively simpler techniques; | o Lesser segmentation accuracy;<br>o Intense computation necessitates high computational power;<br>o Not suited for real time applications; |
| **Dense sliding window methods** | ✓ Relatively unexplored area;<br>✓ Modest segmentation accuracy; | o Use complex algorithms;<br>o Difficult to train and optimize;<br>o Not suited for real time applications; |

## 6. Scope for Future Work

Instance segmentation remains a challenging task. For example, on the popular MS COCO Dataset, the overall average precision is around 50%, leaving plenty of room for improvement. Even as of now, researchers are occupied with the *Hardware Required v/s Algorithmic Simplicity,* as well as *Speed v/s Accuracy* tradeoffs respectively. As an example, in [113] they train the model on an 8-GPU machine where each GPU has 8 clips in a mini-batch (in total a mini-batch size of 64 clips). The models are trained for 400k iterations in total, starting with a learning rate of 0.01 and reducing it 10 times for every 150k iterations.

Hardware constraints limit the scope of research. As can be seen, tasks like instance segmentation are computationally expensive. In spite of this real-time instance segmentation (speed optimization) remains an issue. This has potential applications in intelligent systems like autonomous vehicle systems, security applications, biometrics, etc.

At the model design level, efficient management of the feature *flood* (due to complex architectures) at large, and self-automation of fine-grained convolution metrics like stride for better results, are still issues. Small object detection remains quite challenging. End-to-end based systems design and their training remain issues.

Body part detection research had also generated interest. Datasets on human pose estimation and human parsing (MHPv1.0 [140], MHPv2.0 [141] and Pascal Person Part Database [142]) have been made available, preliminary results [143,140,67,141] are available and new discoveries are around the corner.

## 7. Conclusion

In this paper, an overview of instance segmentation is given. The evolution of image segmentation is from coarse to fine inference. This evolution has come up to instance segmentation and is continuing further, with advancements in computing power and research prowess. In this paper, important instance segmentation issues have been discussed. Various techniques used for instance segmentation have been discussed, from both holistic as well individual perspectives. Their taxonomy, strengths and weaknesses have been discussed. The popular datasets used for instance segmentation has been discussed. Major issues which open the scope for future research have been discussed. The survey is an attempt to provide information about the state of art in the field of instance segmentation, with regards to its purpose, emergence, techniques and related work, datasets and scope for future work.